%% file: main.tex
\documentclass[10pt,twocolumn,letterpaper]{article}

\usepackage[pagenumbers]{cvpr}      
\usepackage{subcaption}

\input{preamble}

\definecolor{cvprblue}{rgb}{0.21,0.49,0.74}
\usepackage[pagebackref,breaklinks,colorlinks,allcolors=cvprblue]{hyperref}

\def\paperID{} 
\def\confName{CVPR}
\def\confYear{2026}

\title{Contact-Aware Neural Dynamics}

\author{
Changwei Jing$^1$, Jai Krishna Bandi$^1$, Jianglong Ye$^1$, Yan Duan$^2$, \\ Pieter Abbeel$^2$,
Xiaolong Wang$^{1\dagger}$, Sha Yi$^{1\dagger}$ \\[6pt]
$^1$UC San Diego, $^2$Amazon FAR (Frontier AI \& Robotics), $^\dagger$Equal advising
}
\begin{document}
\twocolumn[{%
\renewcommand\twocolumn[1][]{#1}%
\maketitle

\begin{center}
  \centering
  \captionsetup{type=figure}
  \includegraphics[width=0.9\textwidth]{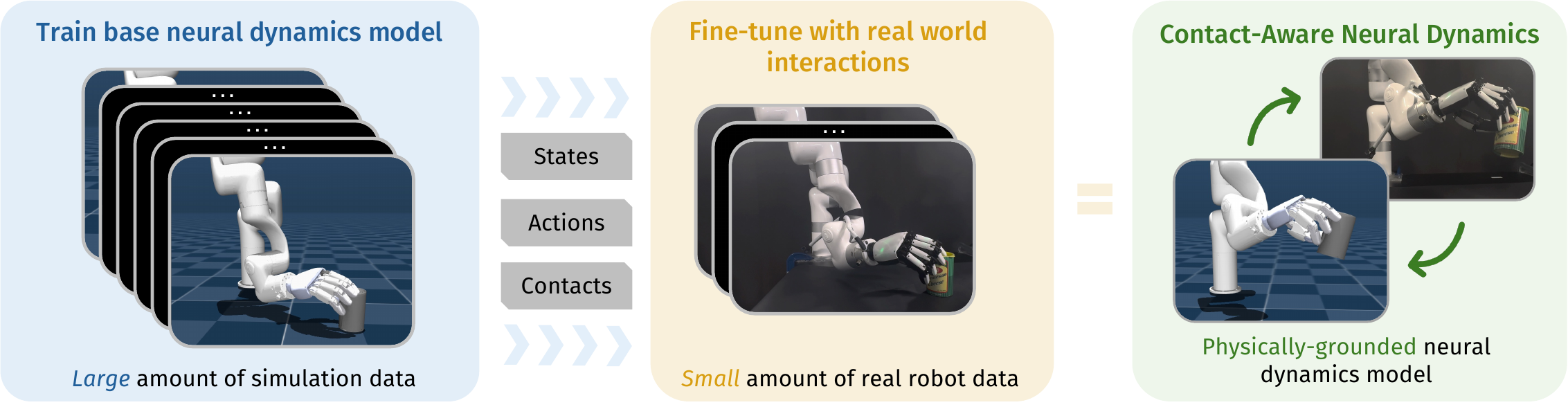}
  \caption{We present an implicit sim-to-real alignment framework for robot manipulation that uses contact information from tactile sensors. We first train a base neural dynamics model using large-scale simulation data, allowing the network to learn contact-induced physical behaviors from histories of states, actions, and contact information. We then fine-tune the model with a small amount of real-world interaction data, including tactile information, enabling it to better capture real contact patterns and dynamics in contact-rich manipulation tasks. This yields a \textbf{contact-aware neural dynamics} model that performs consistently across both simulation and the real world.}
  \label{fig:teaser}
\end{center}
}]

\input{sec/0_abstract}    
\input{sec/1_intro}
\input{sec/2_related_work}
\input{sec/3_method}
\input{sec/4_results}
\input{sec/5_conclusion}
{
    \small
    \bibliographystyle{ieeenat_fullname}
    \bibliography{main}
}


\end{document}

%% file: preamble.tex
\PassOptionsToPackage{table}{xcolor}
\usepackage{xcolor,colortbl,booktabs,siunitx}
\newcolumntype{M}{S[table-format=1.4]}
\newcolumntype{A}{S[table-format=3.2]}



%% file: sec/0_abstract.tex
\begin{abstract}
High-fidelity physics simulation is essential for scalable robotic learning, but the sim-to-real gap persists, especially for tasks involving complex, dynamic, and discontinuous interactions like physical contacts. Explicit system identification, which tunes explicit simulator parameters, is often insufficient to align the intricate, high-dimensional, and state-dependent dynamics of the real world. To overcome this, we propose an implicit sim-to-real alignment framework that learns to directly align the simulator's dynamics with contact information. Our method treats the off-the-shelf simulator as a base prior and learns a contact-aware neural dynamics model to refine simulated states using real-world observations. We show that using tactile contact information from robotic hands can effectively model the non-smooth discontinuities inherent in contact-rich tasks, resulting in a neural dynamics model grounded by real-world data. We demonstrate that this learned forward dynamics model improves state prediction accuracy and can be effectively used to predict policy performance and refine policies trained purely in standard simulators, offering a scalable, data-driven approach to sim-to-real alignment.
\end{abstract}

%% file: sec/1_intro.tex
\section{Introduction}
\label{sec:intro}
Although recent advances in teleoperation and human-in-the-loop data collection have enabled impressive robot demonstrations of dexterous manipulation, simulation-based approaches for training and evaluation provide significantly greater scalability and diversity. Learning purely from real-world teleoperation is expensive, time-consuming, difficult to iterate, and limited to only human intuition. This makes accurate physics simulation crucial for developing generalizable manipulation policies. However, the gap between simulated and real-world dynamics is still big, particularly for contact-rich manipulation tasks. Policies trained in simulation have transferred successfully to a range of locomotion problems, where dynamics are only needed for robots and a static ground, so that modeling errors are often forgiving. Directly deploying sim-trained policies for in-hand or contact-rich manipulation is far less reliable. Small discrepancies in contact geometry, frictional modeling, compliance, or simulation integration timing can drastically change object motion and stability, making standard rigid-body simulators insufficient for faithfully capturing the delicate dynamics for dexterous manipulation.

A dominant way to mitigate this gap is \textit{explicit} system identification: improving geometric parameters~\cite{abou2024physically, chen2024urdformer}, tuning friction and mass~\cite{pfaff2025}, or optimizing a small set of physical parameters so that simulated rollouts match real trajectories~\cite{chen2025learning}. However, this approach fundamentally assumes that a low-dimensional parametric correction is good enough, which is not sufficient for contact-rich manipulation tasks. Many sources of error are high-dimensional (complex contact formulation involving damping and restitution), state-dependent, or simply due to the discrete integration of the simulation. Attempts to compensate purely via domain randomization or parameter sweeps often trade accuracy for robustness and still fail to capture the non-smooth transitions~\cite{Par2021Stiff} that arise from contact. In parallel, vision-focused sim-to-real techniques, such as rendering~\cite{jiang2025gsworld, qureshi2025splatsim} and aggressive domain randomization~\cite{han2025re}, primarily close the perception gap, enabling robust visual policies while leaving the underlying contact dynamics model largely unchanged.

In parallel, recent work has explored \textit{implicit} alignment via learning. These methods include neural residual models atop analytical dynamics~\cite{gao2024sim, zeng2020tossingbot}, differentiable simulators with learned components~\cite{heiden2021neuralsim}, and neural dynamics models trained on simulation~\cite{xu2025neural} or real-world~\cite{zhang2024adapti} data. However, models trained on purely simulation data often lack transferability, while those using only real data are inefficient to collect.
Many of these approaches treat the simulator as a prior and use a neural network to model the residual errors. A significant limitation is that these methods are often contact-agnostic, i.e., treating discontinuities as noise, or rely solely on kinematic and proprioceptive signals. Consequently, they under-utilize one of the richest signals available in manipulation: contact information itself. High-bandwidth tactile sensing, in contrast, provides a fast and responsive signal to guide the modeling process and future dynamics predictions.

In this work, we propose \textit{contact-aware neural dynamics}: an implicit sim-to-real alignment framework that leverages contact information in both simulation and the real world. We first train a neural forward dynamics model in simulation using large-scale rollouts of a dexterous hand interacting with diverse objects under extensive domain randomization. The model conditions on the robot and object states, robot actions, and rendered contact information. This neural dynamics model learns to predict multi-step rollouts for both successful and failed manipulation trajectories. We then collect corresponding real-world trajectories, again including both successes and failures, augmented with tactile sensor readings, and fine-tune the simulation-only model with real-world data. By co-training rather than fitting a separate correction stage, the learned dynamics implicitly align simulated and real-world states in a shared representation based on contact events. This gives a contact-aware forward model that utilizes the diversity and efficiency in simulation while inheriting fidelity from real-world robot data, enabling more accurate robot policy behavior in contact-rich manipulation.

%% file: sec/2_related_work.tex
\section{Related Work}
\label{sec:relatedwork}

Bridging the sim-to-real gap remains central in robotic learning. Simulators enable scalable and safe data collection but diverge from reality due to mismatches in interaction dynamics.  Domain randomization perturbs simulation parameters such as masses and friction to account for the sim2real mismatch \cite{Tobin2017, Peng2018, Andrychowicz2020}, yet small errors in contact parameters often yield unrealistic dynamics.  Complementary approaches either attempt to tune simulator parameters using data collected in the real world \cite{Muratore2021, Chebotar2019, ye2025power} or infer latent physical variables online for policy conditioning \cite{Yu2017UPOSI}.  Recent real-to-sim pipelines generate physics-aware assets or photorealistic reconstructions for policy transfer \cite{pfaff2025, han2025re, Fa2025Bot}.  Broader world-model frameworks integrate generative video modeling with physical scene evolution for control \cite{Mao2025PhysWorld, Zhen2025TesserAct, Agarwal2025Cosmos}.  
Despite progress, most sim2real methods treat contact indirectly—by randomizing few parameters or adapting rigid-body models that struggle with stiff, discontinuous interactions \cite{Par2021Stiff}.  Our work instead learns a dynamics model grounded in tactile feedback, bridging the sim-to-real gap without the need for explicit parameter tuning.

\textbf{System Identification.}
System identification estimates parameters linking simulation and reality.  For manipulation, robot dynamics are usually known, leaving object properties as the main uncertainty.  Classical methods recover mass or inertia from excitation trajectories or force–torque sensing \cite{Khalil2007, Leboutet2021}, but rely on precise calibration.  Automated pipelines such as \cite{pfaff2025} and active exploration frameworks \cite{Kruzliak2024} use robot interaction to infer geometry and inertial parameters, while vision-language models estimate material properties from appearance \cite{Zhai2024FeatureFields}.  Yet these approaches depend on user-specified parameter sets and often fail to generalize when unmodeled effects (compliance or anisotropic friction) dominate.  

\textbf{Vision-Based Methods.}
Vision-based dynamics models aim to infer physics directly from images or videos.  Early methods estimated static object properties such as mass or volume from RGB-D data \cite{Mavrakis2020, Standley2017}, but many quantities (e.g., stiffness or friction) remain visually unobservable.  Deep video-prediction approaches forecast future frames from actions \cite{Oh2015, Finn2016}, and object-centric architectures such as \cite{Chang2017} model pairwise interactions.  Latent world models like \cite{Hafner2019PlaNet, Hafner2020Dreamer} use compact visual dynamics for long-horizon control, while \cite{Ebert2018} applies such models to real manipulation.  More recent works leverage large-scale video and video-language models for planning or imitation \cite{Du2023UniPi, Du2024VLP, Bharadhwaj2024Gen2Act, Jang2025DreamGen, jiang2024robots}.  Although visually coherent, these models often lack physical grounding and predicted trajectories often violate contact realism under occlusion or multi-object interaction.  Physics-aware visual frameworks such as \cite{Mao2025PhysWorld, han2025re} highlight the value of coupling perception with physics; our work follows this principle but grounds learning in tactile rather than visual signals.

\textbf{Neural Dynamics and World Models.}
Neural simulators learn forward dynamics directly from data, offering differentiable, adaptable alternatives to analytical physics replacing analytical contact solvers with learned modules for articulated bodies \cite{xu2025neural}.  Material-conditioned approaches such as \cite{Mittal2024UniPhy, zhang2024adapti} generalize across deformable materials, and particle- or point-cloud world models such as \cite{Huang2024ParticleFormer} capture multi-object interactions.  Hybrid schemes such as \cite{zhang2024adapti} combine particle and grid representations to model deformable-object behavior, while \cite{Jiang2025PhysTwin} reconstructs geometry and physical properties from sparse videos.  Large-scale embodied world models such as \cite{Zhen2025TesserAct, Agarwal2025Cosmos} merge generative vision and physics, pointing toward unified physical AI generation.  However, most rely on full-state or visual supervision and remain weakly grounded in real contact physics.  Our approach complements these efforts by learning a tactile-aware dynamics model that aligns simulated trajectories with real contact behavior.

%% file: sec/3_method.tex
\section{Method} \label{sec:method}
\begin{figure*}[tb]
\centering
\includegraphics[width=0.9\textwidth]{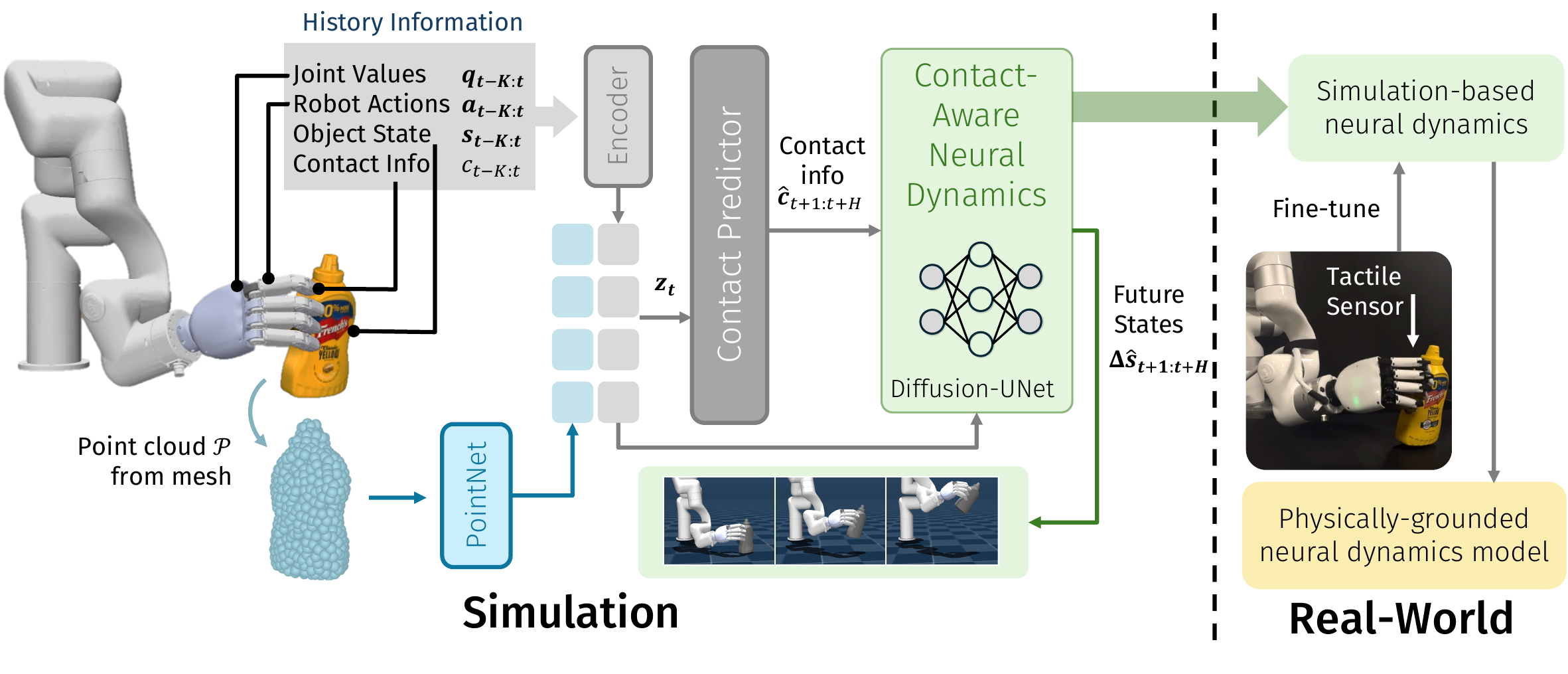}
\caption{\textbf{Overview of the proposed contact-aware neural dynamics framework.}
The model takes as input a multimodal history 
$\mathcal{H}_t=\{\mathbf{s}_{t-K:t},\,\mathbf{q}_{t-K:t},\,\mathbf{a}_{t-K:t},\,c_{t-K:t},\,\mathcal{P}\}$,
including past object poses, joint values, robot actions, binary contact signals, and the object point cloud.
A temporal encoder extracts features from the state--action--contact sequence, while a PointNet encoder processes the geometry $\mathcal{P}$. 
Their fused latent representation $\mathbf{z}_t$ is used by a contact prediction module to infer future contacts $\hat{c}_{t+1:t+H}$, which then condition a diffusion-based pose predictor that outputs future pose increments $\Delta\hat{\mathbf{s}}_{t+1:t+H}$. 
The model is first trained on large-scale simulation data and subsequently fine-tuned with a small amount of real-world interaction data, enabling implicit alignment of simulated and physical contact dynamics.}

\label{fig:method}
\end{figure*}

We formulate sim-to-real alignment for contact-rich manipulation as a \emph{conditional dynamics prediction} problem, where we learn to model contact-dependent object motion using multimodal observations from simulation and reality.
At each time step $t$, the state of the object is defined by its pose $\mathbf{s}_t \in \mathrm{SE}(3)$, represented with translation and rotation. The robot hand joint configuration is denoted as $\mathbf{q}_t \in \mathbb{R}^{d_q}$, and it interacts with the object with its action $\mathbf{a}_t \in \mathbb{R}^{d_q}$. We model the contact between robot hand and object with a binary indication as $c_t \in \{0,1\}$. $c_t = 1$ if \emph{any} fingertip is in contact with the object, and zero otherwise.  
The object geometry is represented by point cloud $\mathcal{P} \in \mathbb{R}^{N\times3}$. The point cloud is obtained by sampling the object mesh surface in the beginning, and transformed by the object pose throughout the entire trajectory. We introduce four more definitions as follows:

\noindent\textbf{History window.}
The input to our neural dynamics model is a sequence of past states and actions. We use a fixed-length observation history over $K{+}1$ past steps:
\begin{equation}
\mathcal{H}_t = \big\{\, \mathbf{s}_{t-K:t},\; \mathbf{a}_{t-K:t},\; \mathbf{q}_{t-K:t},\; c_{t-K:t},\; \mathcal{P} \,\big\},
\label{eq:history}
\end{equation}
which provides geometric, kinematic, control, and contact information for predicting future motion. 
To improve generalization to real-world dynamics from simulation data, we generate trajectories with domain randomization. This includes adding small Gaussian noise to each control command as well as perturbing object and hand poses at random intervals, ensuring that the model is exposed to diverse motion patterns and is robust to actuation noise and state estimation uncertainty.


\noindent\textbf{Prediction goal.}
Given the history $\mathcal{H}_t$, the neural dynamics model aims to predict a future horizon of hand--object contacts and state trajectories:
\begin{equation}
\begin{aligned}
\hat{c}_{t+1:t+H} &= f_{\phi}(\mathcal{H}_t),\\
\Delta\hat{\mathbf{s}}_{t+1:t+H} &= g_{\theta}(\mathcal{H}_t,\, \hat{c}_{t+1:t+H}).
\end{aligned}
\label{eq:targets}
\end{equation}
Here, $f_{\phi}$ denotes the contact-prediction module that infers future binary contact events from the historical observations, while $g_{\theta}$ denotes the state-prediction module that generates future pose trajectories conditioned on both the history and the predicted contact sequence.


\noindent\textbf{Contact representation.}
The contact information in the simulation and real-world may differ greatly, especially when the tactile sensors on the robot hardware are inaccurate and noisy. For simplicity and robustness, we use a \emph{binary, hand-level contact} signal $c_t \in \{0,1\}$. The contact predictor is supervised using a binary cross-entropy (BCE) objective, and the predicted probabilities are re-encoded as a low-dimensional contact feature that conditions the pose dynamics, ensuring contact-aware motion generation.

\noindent\textbf{Neural dynamics representation.}
We represent the system at time $t$ by the object pose $\mathbf{s}_t$, a fixed-length observation history $\mathcal{H}_t$, and the object point cloud $\mathcal{P}$. 
The history $\mathcal{H}_t$ contains object poses $\mathbf{s}_{t-K:t}$, robot actions $\mathbf{a}_{t-K:t}$, joint values $\mathbf{q}_{t-K:t}$, and binary contact states $c_{t-K:t}$. In the next section, we introduce the framework of using this formulation to train a contact-aware neural dynamics model to automatically align simulation and real-world distribution for contact-rich robot manipulation tasks.

\subsection{Model Framework and Architecture}
\label{subsec:architecture}

Our model consists of two coupled modules:
(i) a \emph{Contact Predictor} module that predicts the future contact probabilities $\hat{c}_{t+1:t+H}$, and  
(ii) a \emph{Diffusion Pose Predictor} that generates future pose differences conditioned on the given state and action history, and the predicted contact information.

\noindent\textbf{Multimodal encoders and fusion.}
As shown in  Fig.~\ref{fig:method}, the temporal history information, including the object pose $\mathbf{s}_{t-K:t}$, action $\mathbf{a}_{t-K:t}$, and robot joint configuration $\mathbf{q}_{t-K:t}$, is stacked together along the history dimension as inputs to the neural network. The contact sequences $c_{t-K:t}$ are encoded by an individual module.
The static object point cloud $\mathcal{P}$ is processed by a PointNet encoder~\cite{qi2016pointnet}, yielding a geometry embedding $\mathbf{f}_{\mathcal{P}}$.  
All modality embeddings are concatenated and fused through a lightweight MLP to obtain a shared latent feature $\mathbf{z}_t \in \mathbb{R}^{512}$. This latent feature serves as the input to our two-stage dynamics modeling pipeline, 
where Stage I predicts future contact events and Stage II performs contact-conditioned 
pose diffusion.

\noindent\textbf{Stage I: Contact Predictor.}
Given $\mathbf{z}_t$, an MLP predicts an $H$-step contact probability sequence:
\begin{equation}
\hat{c}_{t+1:t+H} = \sigma(\mathbf{W}_c \mathbf{z}_t + \mathbf{b}_c),
\label{eq:contact_pred}
\end{equation}
\begin{equation}
\mathcal{L}_{\mathrm{cnt}} = \mathrm{BCE}(\hat{c}_{t+1:t+H},\, c_{t+1:t+H}),
\label{eq:contact_loss}
\end{equation}
where $\sigma(\cdot)$ denotes the logistic sigmoid.
The predicted sequence $\hat{c}_{t+1:t+H}$ is further projected through a small MLP 
to obtain a compact contact feature $\mathbf{f}_c \in \mathbb{R}^{d_c}$, 
where $d_c=64$ denotes the dimensionality of the contact embedding vector.

Finally, we form the dynamics condition vector by concatenation:
\begin{equation}
\mathbf{h}_t = [\, \mathbf{z}_t \,;\, \mathbf{f}_c \,].
\end{equation}
Stage~I explicitly augments this latent 
representation with the predicted contact feature $\mathbf{f}_c$, producing the 
contact-conditioned vector $\mathbf{h}_t$ that serves as the input to Stage~II.

\noindent\textbf{Stage II: Diffusion Pose Predictor.}

We model future pose changes with respect to the previous timestep. Let 
$\mathbf{x}_0 = \Delta \mathbf{s}_{t+1:t+H}$ denote the sequence of $H$ pose 
increments, each represented in a 6D minimal form 
$\Delta \mathbf{s}_{t+k} = [\,\Delta \mathbf{p}_{t+k},\, \boldsymbol{\omega}_{t+k}\,]$.
The translation delta is defined as 
\begin{equation}
\Delta \mathbf{p}_{t+k} = \mathbf{p}_{t+k} - \mathbf{p}_{t+k-1},
\label{eq:trans_delta}
\end{equation}
and the rotation increment $\boldsymbol{\omega}_{t+k}\in\mathbb{R}^3$ maps to a 
relative rotation through the exponential map
\begin{equation}
\mathbf{R}_{t+k} 
= \exp(\widehat{\boldsymbol{\omega}}_{t+k})\,\mathbf{R}_{t+k-1},
\label{eq:rotation_update}
\end{equation}
where $\widehat{\boldsymbol{\omega}}$ denotes the skew-symmetric matrix of 
$\boldsymbol{\omega}$.

To model the distribution over $\mathbf{x}_0$, we use a conditional denoising 
diffusion model, whose forward process is
\begin{equation}
q(\mathbf{x}_t \mid \mathbf{x}_0)
= \mathcal{N}\!\big(\sqrt{\bar{\alpha}_t}\,\mathbf{x}_0,\,
(1-\bar{\alpha}_t)\mathbf{I}\big),
\label{eq:forward_diffusion}
\end{equation}
where $\mathbf{x}_t$ is the noisy sample and $\bar{\alpha}_t$ is the cumulative 
noise schedule.

The reverse denoising process is parameterized by a 1D U-Net, introduced here as 
the noise predictor:
\begin{equation}
\boldsymbol{\epsilon}_\theta 
= \mathrm{UNet}_{1\mathrm{D}}(\mathbf{x}_t,\, t,\, \mathbf{h}_t),
\label{eq:unet_predictor}
\end{equation}
with FiLM-modulated conditioning on $\mathbf{h}_t$ applied at all layers.

The training objective is
\begin{equation}
\mathcal{L}_{\mathrm{diff}} =
\mathbb{E}\!\left[\,
\lVert \boldsymbol{\epsilon} -
\boldsymbol{\epsilon}_\theta(\mathbf{x}_t, t, \mathbf{h}_t) \rVert_2^2
\,\right],
\end{equation}
and the trajectory is reconstructed as
\begin{equation}
\hat{\mathbf{s}}_{t+1:t+H} = 
\mathbf{s}_t \oplus \hat{\mathbf{x}}_0,
\end{equation}
where $\oplus$ applies the recovered incremental poses $\hat{\mathbf{x}}_0$ to $\mathbf{s}_t$ over the prediction horizon.

\noindent\textbf{Joint objective.}
The overall training loss combines both stages:
\begin{equation}
\mathcal{L} =
\mathcal{L}_{\mathrm{cnt}} + \lambda\,\mathcal{L}_{\mathrm{diff}},
\label{eq:joint_loss}
\end{equation}
which jointly optimizes contact forecasting and contact-conditioned dynamics, yielding stable, physically consistent, and long-horizon motion predictions.

\subsection{Contact Modeling}
\label{subsec:contact_modeling}
In manipulation tasks, contact dynamics between the fingers and the object are highly non-smooth: force spikes and velocity discontinuities occur at the moment of touch, making continuous contact quantities difficult to model reliably. Instead of regressing continuous contact forces or distributions, we adopt a more robust and learning-friendly \textit{binary, hand-level} contact representation that focuses on the structural signal of whether contact occurs. In practice, continuous contact measurements from tactile or force sensors are noisy, sensitive to calibration, and often exhibit small fluctuations even when the qualitative contact state does not change. Using a binary signal reduces the impact of these fluctuations and gives cleaner supervision for training. Moreover, the binary labels can be consistently derived from both simulation (via collision detection) and real hardware (via tactile sensor readings), which helps align the contact representation across simulation and real-world. This design choice also complements the smooth nature of neural networks: instead of forcing the network to fit high-frequency variations in contact magnitude, it learns to predict the underlying discrete event of contact, which is then used as a stable conditioning signal for the downstream dynamics model.

\noindent\textbf{Simulation.}
In simulation, fingertip and object collision meshes in MuJoCo~\cite{todorov2012mujoco} are used to compute binary contact signals. A contact is labeled as $c_t{=}1$ if any fingertip mesh intersects the object mesh at time $t$, and vice versa.

\noindent\textbf{Real world.}
In real experiments, the XHand fingertips are equipped with force-based tactile sensors, with more detailed information in Section~\ref{subsec:setup}. A fingertip is considered in contact when its measured normal force exceeds a threshold $\tau_{\text{force}}$. The global contact label is set to $c_t{=}1$ if any fingertip surpasses this threshold. This threshold-based definition provides a consistent and robust binary contact representation across simulation and the real system.

\subsection{Implicit Sim-to-Real Alignment with Contacts}
We begin by training our dynamics model entirely in simulation, following the architecture illustrated in Fig.~\ref{fig:method}. The model takes as input the recent history of object states, joint values, robot actions, and binary contact signals, together with the object point cloud. A temporal encoder processes the state–action–contact history, while a PointNet-based geometry encoder extracts shape features from the point cloud. Their outputs jointly condition the denoising diffusion model, which predicts future contact events and object-relative state transitions.

To further enhance the model’s representation of contact-induced dynamics, we perform fine-tuning using simulation data only. This stage leverages a larger and more diverse set of simulated trajectories while maintaining the same contact modeling framework to refine the model. For the single-object setting, we use the mustard bottle from the YCB~\cite{7251504} dataset with 8,000 simulated trajectories. For the multi-object setting, we employ 15,000 simulated trajectories covering 40 YCB objects, with randomized physical and contact parameters to improve robustness and generalization.

All object point clouds are uniformly sampled from their corresponding meshes to ensure consistent geometric representations. Fine-tuning continues from the pretrained weights of the base dynamics model and adopts a lower learning rate to stabilize optimization and further refine the contact-conditioned latent representation. Although no real-world data are used at the first stage, the simulation base model serves as a strong prior, providing a solid foundation for subsequent co-training with real data and improving consistency with tactile contact behaviors observed in real experiments.


%% file: sec/4_results.tex
\section{Results}
We evaluated our contact-aware neural dynamics model in both simulation and real-world settings to assess its capability to bridge the sim-to-real gap and accurately model contact-induced motion in robot manipulation tasks. Our experiments are designed to examine: (i) long-horizon forward prediction accuracy, (ii) generalization across diverse objects and contact configurations, and (iii) the impact of contact conditioning on sim-to-real transfer.

Overall, the results demonstrate that incorporating explicit contact representations significantly enhances both the physical realism and temporal consistency of the predicted trajectories. Compared with previous neural dynamics baselines~\cite{zhang2024particle}, our model achieves lower trajectory prediction errors, more stable multi-step rollouts, and improved alignment with real tactile signals. Qualitatively, it produces physically plausible object motions that closely match observed contact transitions, while quantitatively achieving substantial gains in MSE and ADD-S metrics across both single-object and multi-object scenarios.

These findings highlight that contact-aware learning provides an effective and scalable framework for aligning simulated and real-world dynamics, enabling robust forward prediction and reliable policy evaluation in contact-rich manipulation tasks.

\subsection{Experiment Setup}
\label{subsec:setup}

\begin{figure}[tb]
\centering
\includegraphics[width=\linewidth]{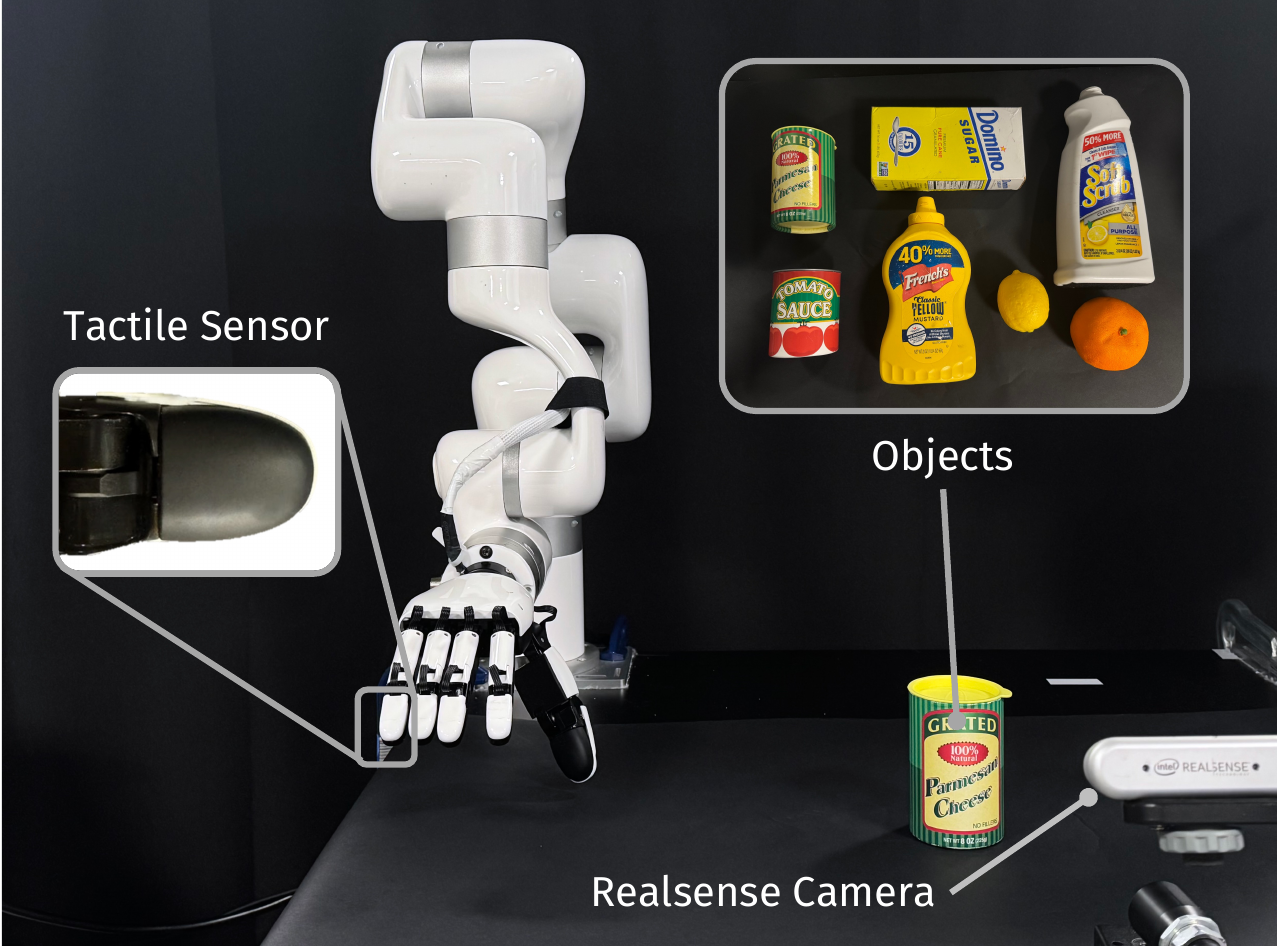}
\caption{Real-world setup with an XArm7 robot arm and an XHand equipped with tactile sensors for contact detection. A Realsense camera captures visual observations, and everyday YCB objects are used for grasping.}
\vspace{-0.3cm}
\label{fig:realworld_setup}
\end{figure}

The real-world setup is shown in Figure~\ref{fig:realworld_setup}. Our system consists of an XArm7 robotic arm equipped with the XHand. A collection of everyday objects is used to evaluate contact-rich grasping and manipulation tasks. Each fingertip of the XHand is equipped with a force-based tactile sensor, and contact is detected when the measured normal force exceeds a predefined threshold, providing reliable binary contact signals during real-world experiments. 

To obtain the real-world object poses, we used FoundationPose \cite{foundationposewen2024} as the pose estimation backbone. To mitigate the residual noise and drift introduced by vision-based estimation, we adopt a lower control frequency and apply small random perturbations to the measured poses during training, which effectively regularizes the dynamics model and reduces the impact of pose errors.

\subsubsection{Tactile Sensor Setup}

The XHand is equipped with five fingertip modules-one per finger - each consisting of an array of tri-axial tactile sensors capable of capturing detailed contact forces along the $x$, $y$, and $z$ axes at around 120 uniformly distributed points on the fingertip with a minimum resolution of 0.05\,N. These tactile arrays enable fine-grained perception of local contact distributions, making them suitable for manipulation tasks requiring precise force feedback.
During operation, the XHand communicates via an RS485 interface with the control and sensing loop running at approximately 80–85\,Hz.
For each fingertip sensor, the high-level computed force vector ${F}_{\text{calc}} = [F_x,\, F_y,\, F_z]$,
    representing the aggregated 3D contact force at the fingertip is measured.

Before data collection, all sensors undergo a reset and calibration routine to remove static offsets and ensure consistent baselines. The calibration procedure records the mean force over a stationary 3\,s window and subtracts it from subsequent readings:

\begin{equation}
    \mathbf{F}_{\text{calibrated}} = \mathbf{F}_{\text{calc}} - \mathbf{F}_{\text{offset}}.
\end{equation}

A lightweight contact detection heuristic identifies contact events based on the cumulative force magnitude:
\begin{equation}
c_i =
\begin{cases}
1, & \text{if } |F_x| + |F_y| + |F_z| > 0.3N,\\[4pt]
0, & \text{otherwise,}
\end{cases}
\end{equation}
yielding a binary contact flag $c_i$ for each fingertip.

\subsection{Qualitative Results}
\begin{figure*}[tb]
  \centering
  \includegraphics[width=\linewidth]{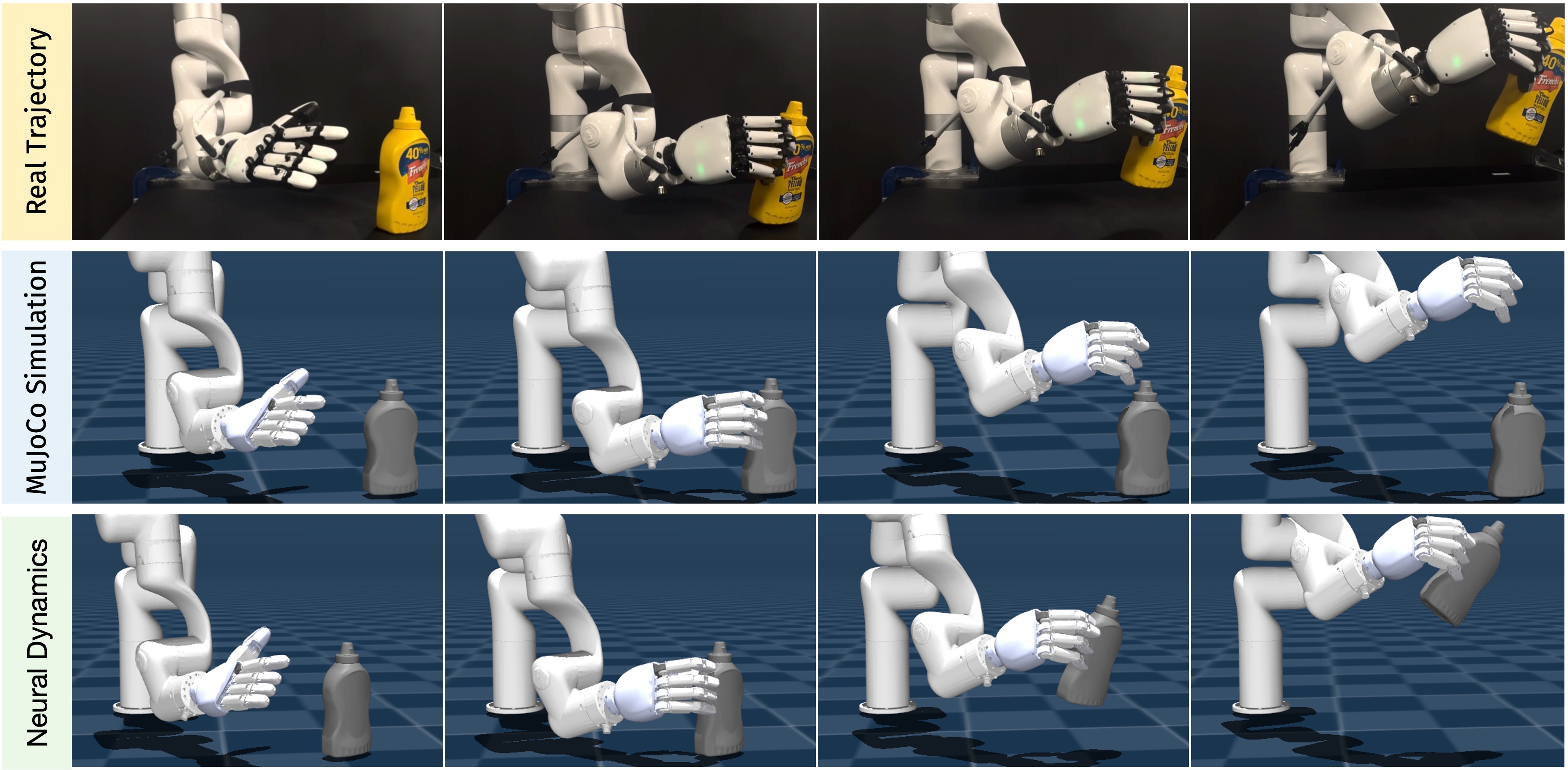}
  \vspace{-1.0em}
  \caption{
    \textbf{Qualitative comparison between real, simulated, and our contact-aware neural dynamics results.}
The first row shows real-world rollouts, the second shows standard MuJoCo simulations, and the third shows our model predictions. Standard simulation often yields unstable or incorrect contacts and physically implausible object motion due to limited contact modeling and the sim-to-real gap. In contrast, our model produces smoother, more realistic trajectories aligned with real-world motion. When co-trained with a small amount of real data, it further improves temporal stability and contact consistency, demonstrating stronger sim-to-real transfer.
  }
  \label{fig:real_sim_ours}
  \vspace{-0.2em}
\end{figure*}

We present qualitative evaluations to illustrate how our contact-aware neural dynamics model bridges the sim-to-real gap and improves long-horizon prediction fidelity.

\begin{figure}[tb]
  \centering\includegraphics[width=\linewidth]{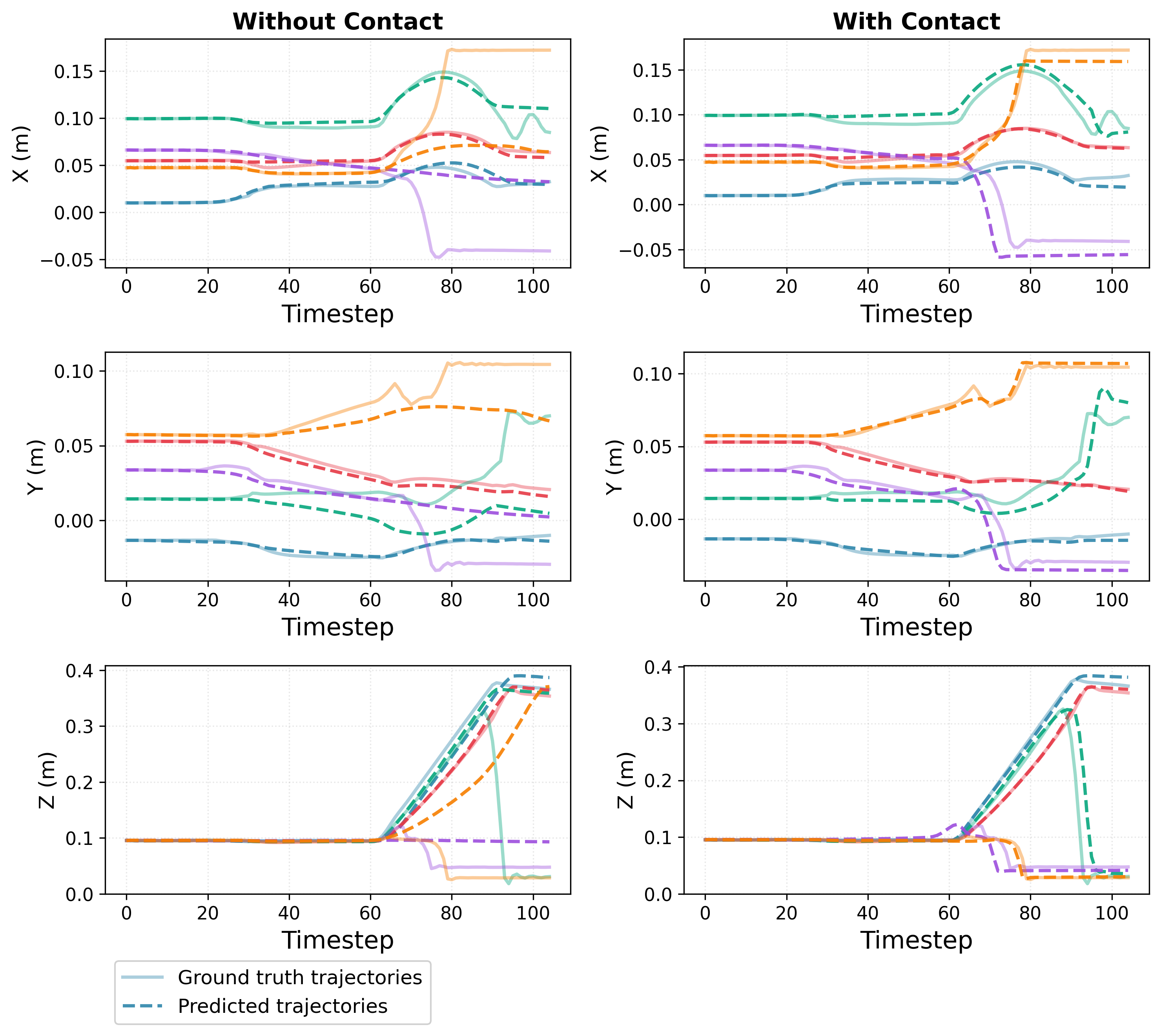}
  \vspace{-0.8em}
  \caption{Comparison of multi-trajectory rollouts: the left panel shows predictions from our two-stage contact-aware model versus ground truth, while the right panel shows rollouts from a direct dynamics predictor compared with the same ground truth.}
  \label{fig:rollout_multi}
  \vspace{-0.5em}
\end{figure}

As qualitatively demonstrated in Fig.~\ref{fig:real_sim_ours}, our method produces simulated rollouts exhibiting high fidelity to real-world observations. It faithfully captures object motion and abrupt contact transitions---nuances standard simulators often miss due to simplified dynamics. Even under significant variations in contact geometry, our model preserves structural integrity and generates physically plausible states, highlighting robustness against domain randomization and parameter mismatch.

Figure~\ref{fig:rollout_multi} compares multi-step 3D trajectory rollouts between ground truth and predictions. The proposed two-stage architecture is critical: by conditioning diffusion dynamics on inferred contact cues, it maintains spatial consistency and models contact-driven discontinuities. In contrast, single-step predictors neglecting contact suffer from compounding errors, leading to significant drift. Our model adaptively switches motion regimes, resembling a simulator with collision detection. This capability is evident in slippage scenarios, where the model detects contact loss and adjusts the trajectory, yielding stable, physically consistent predictions.

Overall, these results validate that integrating contact-aware representations bridges the sim-to-real gap. Our approach generalizes effectively, ensuring synthesized predictions remain physically grounded and temporally coherent in contact-rich manipulation tasks.

\subsection{Performance Comparisons}
\input{table/compare_results}
We further provide quantitative comparisons in Table~\ref{tab:single_multi}, evaluating different dynamics models under both single- and multi-object settings across three data regimes (simulation, real, and co-training). Performance is measured by Mean Squared Error (MSE) and the \textbf{area under the curve (AUC) of ADD-S}~\cite{xiang2018posecnn}. Hereafter, we simply refer to this metric as \textbf{ADD-S}. Higher is better.
 Lower MSE and higher ADD-S indicate better prediction accuracy. Intuitively, \textbf{ADD-S} captures how often the predicted and ground-truth trajectories stay within distance thresholds in 3D space, reflecting how well the model preserves the object’s geometric consistency during long-horizon motion prediction. As shown in the table, our \textit{Diffusion-UNet with contact} achieves the best performance across all regimes, consistently yielding lower prediction errors and higher spatial accuracy.

In particular, the model exhibits significant gains under the co-training setup, which achieves \textbf{0.0082~MSE} and \textbf{88.23\%~ADD-S} in single-object tasks, while also maintaining strong performance in the more complex multi-object setting. 
These results highlight that incorporating contact-aware representations not only improves physical realism but also enhances sim-to-real transfer and generalization.

\subsection{Applications}
Neural forward dynamics models have shown broad applicability in capturing complex physical interactions. Recent works~\cite{liu2025dexndmclosingrealitygap, Hansen2022tdmpc} demonstrate that such forward models can not only accurately predict future object motion, but also infer latent physical properties and provide differentiable structure for downstream decision-making. By learning contact-induced dynamics directly from data, these models can replace or augment traditional simulators, enabling more precise dynamics prediction and control in challenging manipulation scenarios. Beyond prediction, neural forward models serve as strong priors for control and policy learning: they can be integrated into model-based planning frameworks or used to fine-tune purely simulation-trained policies (e.g., large-scale dexterous manipulation policies such as Dex1B \cite{ye2025dex1b}), allowing better adaptation to real-world friction, compliance, and contact patterns. Overall, neural dynamics models offer a flexible and unified foundation for simulation, inference, and control, and hold significant promise for contact-rich robotic manipulation. An example application we performed is to use this neural physics model to evaluate and filter a manipulation policy trained only in simulation.

\input{table/success_table}

\noindent\textbf{Task Success Rate Evaluation.}
We evaluated our neural dynamics model using the \emph{task success rate}, 
defined as the percentage of rollouts whose final predicted object position deviates 
from the real-world trajectory endpoint by less than \SI{5}{cm}. This metric 
quantifies whether the model can maintain accurate and drift-free long-horizon 
predictions that remain consistent with real object motion.

We evaluate two training regimes: (1) \textbf{Real-only}, trained solely on 
real-world trajectories; and (2) \textbf{Sim+Real w/ Contact}, which is pretrained 
on large-scale simulation data and subsequently fine-tuned using real-world 
contact observations. As shown in Table~\ref{tab:task_success_rate}, the 
Sim+Real w/ Contact model achieves significantly higher success rates in both 
single-object and multi-object scenarios, reaching \textbf{73.7\%} and 
\textbf{64.7\%}, respectively. In contrast, the Real-only model suffers from 
accumulated prediction drift over long horizons. Leveraging structured priors 
from simulation and refining them with real contact supervision enables our 
model to effectively bridge the sim-to-real gap.

These results demonstrate that incorporating contact-aware dynamics not only 
improves physical prediction fidelity but also leads to substantially higher 
robustness and success in downstream manipulation tasks.

%% file: table/compare_results.tex
\begin{table*}[tb]
\centering
\caption{Quantitative comparison of single-object and multi-object tasks under different training regimes (simulation only and simulation+real co-training). Metrics: mean squared error (MSE$\downarrow$) and \textbf{AUC of ADD-S} (\%\,$\uparrow$). Hereafter we abbreviate it as \textbf{ADD-S}. Our model outperforms neural dynamics baselines in both settings, with further gains after limited real-world fine-tuning, especially for multi-object tasks, highlighting the strong generalization of our contact-aware representation and its ability to reduce the sim-to-real gap.}

\label{tab:single_multi}
\small
\setlength{\tabcolsep}{4.2pt}
\renewcommand{\arraystretch}{1.12}
\begin{tabular}{@{}l *{12}{c}@{}}
\toprule
& \multicolumn{6}{c}{\textbf{Single object}} & \multicolumn{6}{c}{\textbf{Multiple objects}} \\
\cmidrule(lr){2-7}\cmidrule(lr){8-13}
& \multicolumn{2}{c}{Sim data} & \multicolumn{2}{c}{Real data} & \multicolumn{2}{c}{Real-Finetune}
& \multicolumn{2}{c}{Sim data} & \multicolumn{2}{c}{Real data} & \multicolumn{2}{c}{Real-Finetune} \\
\cmidrule(lr){2-3}\cmidrule(lr){4-5}\cmidrule(lr){6-7}
\cmidrule(lr){8-9}\cmidrule(lr){10-11}\cmidrule(lr){12-13}
{Method}
& {MSE$\downarrow$} & {ADD-S$\uparrow$}
& {MSE} & {ADD-S}
& {MSE} & {ADD-S}
& {MSE} & {ADD-S}
& {MSE} & {ADD-S}
& {MSE} & {ADD-S} \\
\midrule

\rowcolor{white}
Baseline~\cite{zhang2024particle} & 0.016 & 71.01 & 0.0194 & 68.72 & --- & --- & 0.0159 & 61.60 & 0.0161 & 62.98 & --- & --- \\

\rowcolor{gray!15}
MLP & 0.026 & 62.58 & 0.0130 & 78.12 & 0.0110 & 77.43 & 0.0150 & 60.11 & 0.0082 & 71.76 & 0.0069 & 73.43 \\

\rowcolor{white}
UNet & 0.022 & 65.86 & 0.0150 & 68.45 & 0.0130 & 70.11 & 0.0170 & 67.74 & 0.0084 & 72.09 & 0.0085 & 74.12 \\

\rowcolor{gray!15}
Diffusion-UNet & 0.021 & 69.12 & 0.0098 & 80.03 & 0.0091 & 82.45 & 0.0120 & 69.95 & 0.0083 & 73.04 & 0.0065 & 75.82 \\

\rowcolor{white}
Diffusion-UNet w/ Contact & 0.015 & 68.12 & 0.0094 & 81.34 & \textbf{0.0082} & \textbf{88.23} & 0.0100 & 69.34 & 0.0075 & 73.33 & \textbf{0.0058} & \textbf{79.12} \\

\bottomrule
\end{tabular}
\end{table*}


%% file: table/success_table.tex
\begin{table}[t]
    \centering
    \caption{Task success rate under single-object and multi-object settings. 
    Success is defined as the percentage of rollouts whose final predicted 
    object position deviates from the real trajectory endpoint by less than 
    \SI{5}{cm}.}
    \label{tab:task_success_rate}
    \begin{tabular}{lcc}
        \toprule
        \textbf{Method} & \textbf{Single-object (\%)} & \textbf{Multi-object (\%)} \\
        \midrule
        Real-only & 52.6 & 47.1 \\
        Sim+Real & 73.7 & 64.7 \\
        \bottomrule
    \end{tabular}
\end{table}

%% file: sec/5_conclusion.tex
\section{Conclusions and Limitations}
This paper presents a contact-aware neural dynamics model for contact-rich manipulation tasks. By combining a unified binary contact representation, a geometry-aware point cloud encoder, and a conditional diffusion structure, our framework enables stable prediction of future object motion across multi-object and multi-contact scenarios. Large-scale simulation data provide a robust prior for contact-induced dynamics, while limited real-world co-training further improves the model's adaptability to real friction, sensing noise, and non-ideal contact behaviors. Experimental results demonstrate that our method significantly outperforms existing neural dynamics baselines in both single-object and multi-object settings, effectively narrowing the sim-to-real gap and offering a promising direction for building generalizable manipulation models.

Despite these strengths, our approach has several limitations. First, the model relies on object states estimated by FoundationPose during data collection, whose accuracy may degrade under occlusion, clutter, or multi-object stacking, leading to accumulated errors during prediction. Second, while binary contact signals are stable and easy to learn, they cannot fully capture richer real-world contact attributes such as contact area, slip direction, or force distribution. Third, achieving broad generalization across diverse motions and manipulation tasks requires a large and varied dataset, which limits scalability when data collection is costly or task coverage is limited. Finally, although the model performs well over short prediction horizons, long-horizon rollouts still suffer from compounding errors, particularly under frequent contact switching or rapid object motion, constraining its applicability in long-term planning or large-span dynamic prediction. Future work may incorporate more accurate tracking, richer contact representations, and more data-efficient or structured dynamics models to improve long-horizon stability and generalization.